\pdfoutput=1

\documentclass[11pt]{article}

\usepackage[]{acl}

\usepackage{times}
\usepackage{latexsym}

\usepackage[T1]{fontenc}

\usepackage[utf8]{inputenc}

\usepackage{microtype}
\usepackage{amsmath}
\usepackage{amssymb}
\usepackage{todonotes}
\usepackage{csquotes}
\usepackage{listings}
\usepackage{multirow}
\usepackage{booktabs}
\usepackage{graphicx}
\usepackage{array}
\usepackage{pbox}

\newcommand{\causalselfattention}{\mathrm{CausalSelfAttention}}
\newcommand{\attend}{\mathrm{MultiheadAttention}}

\title{ED2LM: Encoder-Decoder to Language Model \\ for Faster Document Re-ranking Inference}
\author{
\textbf{Kai Hui}\thanks{~~Corresponding Author} \quad
\textbf{Honglei Zhuang} \quad
\textbf{Tao Chen} \quad 
\textbf{Zhen Qin}  \quad 
\textbf{Jing Lu} \quad 
\textbf{Dara Bahri}  \quad \\
\textbf{Ji Ma} \quad
\textbf{Jai Gupta} \quad 
\textbf{Cicero Nogueira dos Santos} \quad
\textbf{Yi Tay} \quad  
\textbf{Donald Metzler} \quad  \\\\ kaihuibj@google.com \\ Google Research }

\begin{document}
\maketitle
\begin{abstract}
State-of-the-art neural models typically encode document-query pairs using cross-attention for re-ranking. To this end, models generally utilize an encoder-only (like BERT) paradigm or an encoder-decoder (like T5) approach. These paradigms, however, are not without flaws, i.e., running the model on all query-document pairs at inference-time incurs a significant computational cost. This paper proposes a new training and inference paradigm for re-ranking. We propose to finetune a pretrained encoder-decoder model using in the form of document to query generation. 
Subsequently, we show that this encoder-decoder architecture can be decomposed into a decoder-only language model during inference. 
This results in significant inference time speedups since the decoder-only architecture only needs to learn to interpret static encoder embeddings during inference. Our experiments show that this new paradigm achieves results that are comparable to the more expensive cross-attention ranking approaches while being up to 6.8X faster. We believe this work paves the way for more efficient neural rankers that leverage large pretrained models.
\end{abstract}

\section{Introduction}\label{sec.introduction}

Leveraging transformer architecture to model the concatenation of a query-document pair is a well-established approach for document ranking \citep{nogueira2020T5ranking}.  Today, modern neural methods for re-ranking are based on the encoder-only (e.g., BERT \citep{devlin2019bert}) or encoder-decoder  (e.g., T5 \citep{raffel2020exploring}) paradigm where query-document interactions are modeled by the encoder's attention mechanism. Unfortunately, these paradigms are computationally prohibitive given that the model has to be run on all document-query pairs during inference. To this end, it is commonplace to use less powerful but computationally lightweight dual encoder models \citep{nogueira2019docT5query,karpukhin2020dense,xiong2020approximate,qu2021rocketqa, gao2021coil} for first-pass retrieval and to only run the more expensive re-ranker on a small subset of retrieved candidates. Even with this setup, cross-attention-based re-ranking can still be expensive, especially when larger pretrained Transformer models are used. As such, this paper is primarily concerned with improving inference-time re-ranking efficiency while maintaining comparable effectiveness to existing cross-attention models.

The novelty of this paper lies in a new paradigm for re-ranking that provides up to $6.8$X speedup without any degradation in shallow-pool effectiveness.  Concretely, we propose a new method for inference-time decomposition of encoder-decoder architectures into decoder-only language models. Given a pretrained sequence-to-sequence model, we finetune the encoder-decoder model using a document-to-query multi-task loss. 
At inference, we decompose the encoder-decoder architecture into a decoder-only language model (LM) that learns to interpret from a memory store of encoded document tokens representations using attention. The document-query pair score can be interpreted as the likelihood of generating the query given the encoded document term representations. 

There are multiple efficiency benefits to our proposed design. First, significant inference-time cost savings are unlocked since the document term memory store can be pre-computed in advance and acts as a read-only memory. Second, our redesign also exploits the fact that queries are generally much shorter than documents. During inference time, only query tokens have to be passed through the decoder stack when attending to the pre-computed document representations which allows us to also obtain an additional speed advantage over encoder-only BERT-like models. Third, computing the query likelihood is computationally simple and does not require the typical costs associated with autoregressive generation models.

The overall contributions of this work can be summarized as follows:
\begin{itemize}
    \item We propose a new re-ranking paradigm, ED2LM (Encoder-Decoder to Language Model) for fast and efficient inference-time re-ranking. Our method is based on inference-time decomposition of an encoder-decoder model into a decoder-only language model.
    \item The proposed method utilizes a new fine-tuning paradigm by incorporating a new objective function that combines the generative query likelihood and the discriminative cross-entropy loss. 
    \item Via extensive experiments, we show that the proposed method performs competitively with T5-based cross-attention re-rankers~\citep{nogueira2020T5ranking}
    while being up to more than 6.8X faster during inference. 
\end{itemize}
\section{Related Work}\label{sec.relatedwork}

\paragraph{Neural text ranking.}
Traditional ranking systems focus on numeric input features~\cite{dasalc, diversification}. Recently, text ranking is popular given the prevalence of large pretrained language models. A number of so-called cross-attention models concatenate a query and a candidate document into a string and feed it into the model~\cite{tfranking,nogueira2020T5ranking,chen2021cobert}, which allows the attention mechanism of the model to capture interactions across query and document terms. 
However, deploying such models to millions or billions of documents is usually intractable due to the exorbitant computational cost. 
To combat this cost, other studies have explored more efficient models, e.g., dual-encoder models~\cite{karpukhin2020dense,qu2021rocketqa,ren2021rocketqav2}, BERT with late interaction~\cite{khattab2020colbert}, or using contextual language models to improve term weighting in traditional inverted indexes~\cite{nogueira2019docT5query, dai2020context,gao2021coil}. 

A few studies that are most closely related to this work focus on leveraging the generative nature of pretrained encoder-decoder language models. 
A natural practice is to directly use the likelihood of generating the query given a document to rank the documents~\cite{Zhuang2021-dn,zhuang2021deep,lesota2021modern}. 
However, these methods mostly perform substantially worse than cross-attention ranking models.
Another work~\cite{dos2020beyond} transforms the likelihood of generating the query into a discriminative loss, where an ``unlikelihood'' loss is introduced for negative query-document pairs. 
Despite relatively better performance than using vanilla maximum likelihood estimation (MLE), we found that their method still underperforms cross-attention ranking models. 
Our proposed method uses a combination of query generation loss and a cross-entropy loss on a specific token, which is capable of achieving comparable performance to cross-attention models. 

\cite{ju2021text} proposes query generation as an auxiliary task during training and shows improved performance. 
However, the proposed model still takes both a query and a document as input in the main ranking task and hence would be as costly as cross-attention ranking models during inference.
Finally, the recent differentiable search index \citep{tay2022transformer} proposes end-to-end ranking via text generation using an encoder-decoder T5 model.

\paragraph{Efficient neural IR.}
Due to the excessive computational cost of inference in pretrained language models, there is a series of studies aiming to improve the efficiency. 

A major trend is to distill expensive models into cheaper ones~\cite{hinton2015distilling,sanh2019distilbert}.
Some distillation approaches have specifically focused on text ranking applications~\cite{zhang2020query,ensemble_distill_ictir, chen2021tinybert, hofstaetter2021improving}.

Another trend is to improve model efficiency by modifying the model architecture.
A typical approach used by 
ColBERT~\cite{khattab2020colbert} and PreTTR~\cite{macavaney2020efficient} defer query-document interactions to upper layers so that part of the model can be pre-computed. 
Our model can be categorized into this class of models, 
except that the late interaction is naturally aligned with the decomposition of encoder-decoder models.
This alignment allows us to better leverage knowledge learned by the model during pretraining, and can be the reason behind our stronger performance compared to ColBERT and PreTTR.

There are a couple of other efficient model structures, 
such as 
early exiting~\cite{soldaini2020cascade,xin2020early}, 
Transformer-Kernel (TK) model~\cite{hofstatter2020interpretable},
and contextualized offline relevance weighting~\cite{chen2021contextualized}.
In terms of storage cost, \citet{cohen2021sdr} proposed the succinct document representation which reduces the dimension of token representation to compress document representations.
These techniques are orthogonal to our study and can be combined with our work to further improve the time and storage efficiency.

\section{The Proposed Method}\label{sec.method}

\begin{figure}[t]
    \centering
    \includegraphics[trim={4cm 0 3cm 0},clip,width=8.5cm]{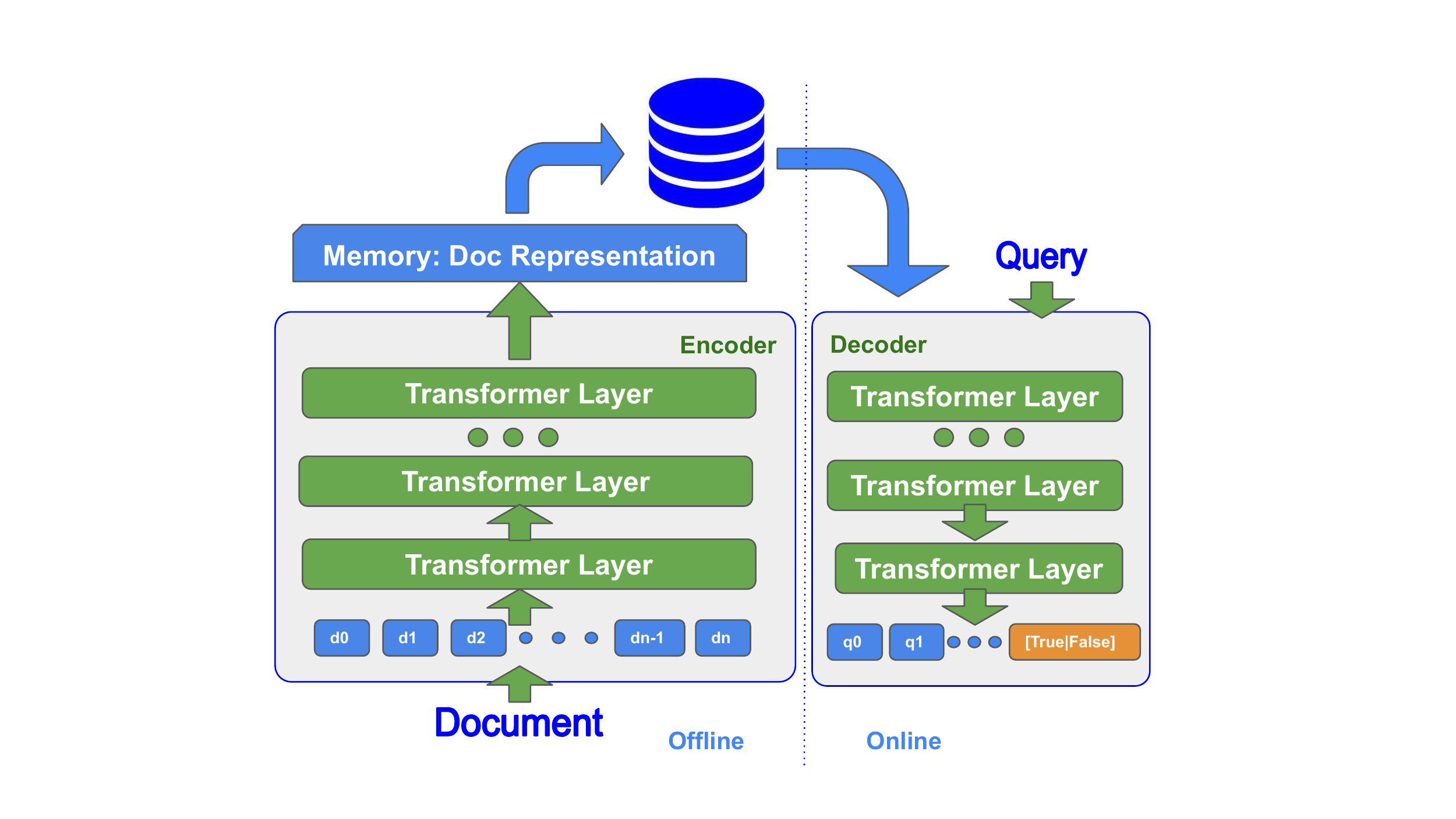}
    \caption{Overview of the proposed ED2LM.}
    \label{fig.method}
\end{figure}

This section describes the ED2LM model. See Fig.~\ref{fig.method} for an overview of the approach.

\subsection{Overview}\label{sec.overview}
The proposed ED2LM model is based on the T5 encoder-decoder architecture. It encodes the documents without looking at the queries and produces ranking scores by decoding the queries and attending to the document representations.

In particular, for a query-document pair, 
the document tokens are encoded with a stack of Transformer layers as in BERT~\cite{devlin2019bert},
where the tokens
attend to one another before going through the position-wise feed-forward layer. 
The output of the encoder is in the form of dense representations for the document tokens.
During decoding, the query tokens are decoded with a stack of decoder layers, 
where 
the query tokens first attend to other query tokens before going through
a multi-head attention block to attend to the document tokens from the encoder.

Inspired by T5~\cite{nogueira2020T5ranking} for ranking and  the use of BART for discrimination~\cite{dos2020beyond,lewis2020bart}, a special true/false token is appended to the end of the query before the end of the query sequence (EOS). During training, inspired by~\cite{ju2021text}, the model is trained to generate the query tokens and determine the relevance of the query-document pair. During inference, only the score for the true/false token is used for ranking.

\subsection{ED2LM for Re-ranking}\label{sec.model}
In this section, we describe the details of training and inference for ED2LM.

\subsubsection{Fine-tuning}
During fine-tuning,  ED2LM involves an encoder-decoder architecture which maps $\mathbb{R}^{L_{D}}$ discrete symbols to $\mathbb{R}^{L_{Q}}$ discrete symbols. Here, $L_{D}$ refers to the length of the document and $L_{Q}$ refers to the query length.

\paragraph{Task formulation.}
The input to the model is a sequence of document tokens and the output of the model is a sequence of query tokens. In order to imbue our model with discriminative capabilities, we append the class token (true/false) that represents the query-document pair at the end of the query. 
The ranking score of a query-document pair 
is the normalised probability of the true token at the end of the query. Given a query $q$ and a document $d$, the ground-truth correctness of $d$ relative to $q$ is denoted as a binary label $y$.

\paragraph{Loss function.} The loss function optimized for  fine-tuning has two components.
The first component is the maximum likelihood estimation (MLE) loss of the individual question tokens, which is defined as:
\begin{equation}\label{eq.loss_ql}
\mathit{Loss_{QL}} = -\sum_{i\in 0\cdots L_{Q}-1}\mathit{log}(P(q_i|q_{:i}; d))
\end{equation}
Since we want the model to learn the correctness of the question using the trailing true/false tokens, we also compute the likelihood of those tokens as follows.
$$\mathit{p}^+=P(\text{true,eos}|q; d)$$
$$\mathit{p}^-=P(\text{false,eos}|q; d)$$
The cross-entropy loss $\mathit{Loss_{CE}}$ can then be written as:
\begin{equation}\label{eq.loss_disc}
\mathit{Loss_{CE}} = -y\mathit{log}\mathit{p}^+ - (1-y)\mathit{log}\mathit{p}^-
\end{equation}
The final training loss can the be written as:
\begin{equation}\label{eq.loss}
\mathit{Loss} = \mathit{Loss_{CE}} + y \mathit{Loss_{QL}}
\end{equation}
The cross-entropy loss is applied to all examples whereas the query likelihood loss only applies to the positive examples. Our fine-tuning loss is trained with teacher forcing.

\paragraph{Scoring.}
The normalised scores from the true and false tokens are combined as
in~\cite{nogueira2020T5ranking}.

\subsection{Efficient Re-ranker}\label{sec.reranker}
This section discusses using ED2LM for more efficient inference, by decoupling the
encoder-decoder into a decoder-only language model. 

\subsubsection{Decomposing Encoder-Decoder to Decoder-only LM}
The key idea for fast inference is to only extract the decoder from the trained Encoder-Decoder model. Recall a decoder-stack is comprised of decoder-side causal self-attention and encoder-decoder cross-attention. 
\begin{align}
X'_{\ell} = \causalselfattention(X_{\ell}, X_{\ell}) \\
Y_{\ell} = \attend(M_{\ell}, X'_{\ell})    
\end{align}
where $X \in \mathbb{R}^{L_{Q} \times d_{model}}$ is the input to the decoder stack at layer $\ell$. $M$ refers to a sequence of memory tokens. In this case, we note that $M$ here refers to computed encoder representations that pass through the encoder-stack. During fine-tuning, this encoder-stack is trained end-to-end. However, this paradigm generalizes these embeddings as \textit{``memory''}, which can be extended to other use cases or applications. We can also interpret this memory as a form of soft prompt.
\subsubsection{Reading from Memory}
The decoder reads from $M$. In the standard setup, $M$ are static representations that originate from the final output of the encoder in the Seq2Seq architecture and the $\attend$ is the encoder-decoder cross attention. Here, $M$ can be compressed 
along the presentation dimension ($d_{model}$) as in~\cite{macavaney2020efficient, gao2021coil, cohen2021sdr}, which is orthogonal to our studies,
or along the sequence dimension ($L_{D}$), which is introduced below.
We find that this generalization is a practically useful way to interpret the ED2LM architecture. 
We propose to explore not only standard $M$ from encoder outputs but also compressed memory stores from Funnel Transformers~\citep{dai2020funnel}.
Herein, we employ the Funnel Transformer with 
$b$ blocks in the encoder, leading to 
$2^b$ storage compression, by 
reducing the $\mathbb{R}^{L_{D}}$ for $2^b$. Between each block, a mean-pooling layer is used to down-sample the input sequence by two in the sequence length
dimension. 

\section{Experiment Setup}\label{sec.setup}

This section describes our experimental setup.

\paragraph{Dataset and metrics.}
We employ the MS MARCO~\cite{nguyen2016ms} passage re-ranking task, for which we report the official
evaluation metric MRR@10 on the 6980 development queries using the binary labels from the dev dataset. We also use the 43 test queries from the TREC Deep Learning (DL)
Track 2019~\cite{craswell2020overview} and the 54 test queries from 2020~\cite{craswell2021overview}.
The TREC data sets include graded relevance judgments. 
We report the official evaluation metrics NDCG@10 as well as mean average precision (MAP).
When computing MAP, following the official TREC setup, we map passage judgments 2 and 3 to relevant and 0 and 1 to non-relevant.
Statistical significance is reported using a paired two-tailed t-test.
We use a maximum sequence length of 256 tokens for paragraphs and 32 tokens for queries in our experiments,
similar to~\cite{hofstatter2020interpretable,hofstaetter2021improving}.

We employ the training data from 
RocketQA~\cite{qu2021rocketqa}, which is derived from the
MS MARCO training dataset as dual-encoder models trained on it demonstrate strong performance.
Specifically, 
we use the hard-question split (``RQA-Hard''), 
which only includes the hard-negative samples and positive samples from MS MARCO, and
the merge split (``RQA-Merge''),
which includes extra unlabeled questions from  
Yahoo! Answers\footnote{\url{http://answers.yahhoo.com}}, ORCAS~\cite{fisch2019mrqa}, and Natural Questions~\cite{kwiatkowski2019natural} on top of 
``RQA-Hard''.
For validation purposes, we use the 1500 dev2 validation queries with at least one relevance judgment from the TREC DL Track 2021\footnote{\url{https://msmarco.blob.core.windows.net/msmarcoranking/passv2_dev2_queries.tsv}}. Given our focus on shallow-pool effectiveness,  the model with highest MRR@10 on the validation dataset is selected.
We employ Mesh Tensorflow~\cite{shazeer2018mesh}
for training and evaluation.
The T5 models have been trained and inferred as 
in~\cite{nogueira2020T5ranking}, and ED2LM has been primarily
trained using the loss defined in Eq.~\ref{eq.loss}. We train models
for ablation study by using Eq.~\ref{eq.loss_ql}
and Eq.~\ref{eq.loss_disc} separately. 
During training, a constant learning rate of $1e\text{-}3$ is used.

\paragraph{Baselines.}
ED2LM is compared to ranking models using four variants of T5 (T5-small, T5-base, T5-large, and T5-xl), BERT-base, BERT-large, and
PreTTR~\cite{macavaney2020efficient}.
The PreTTR~\cite{macavaney2020efficient} model
decouples the encoding of the query and the document on top of the BERT architecture
and is directly comparable to the T5-based ED2LM.
We fine-tune BERT-base 
models using TF-ranking~\cite{TensorflowRankingKDD2019} and achieve similar results with the results reported in~\cite{nogueira2020T5ranking}.
We also re-implement the PreTTR model using TF-ranking. Therein, following the configurations in~\cite{macavaney2020efficient}, a query and a document are encoded independently in the first $l$-layers using the BERT-base configuration before interacting via cross-attention. The BERT-base pre-trained checkpoint is used for initialisation. We report the results by setting $l=6$, which leads to similar FLOPs and latency as ED2LM-base (26.1T vs 20.6T).

\paragraph{Variants of ED2LM.}
We investigate the effectiveness and inference efficiency of ED2LM based on T5-small, T5-base, T5-large, and T5-xl architectures,
leading to ED2LM-small, ED2LM-base, ED2LM-large, and ED2LM-xl, respectively.
We experiment with two Funnel-Transformer variants,  where
 two six-layers funnel blocks ($b=2$)
and three eight-layers funnel blocks ($b=3$) are used
in the encoder, respectively. They are named
ED2LM-F-$6L^{\times2}$
and ED2LM-F-$8L^{\times3}$, correspondingly.
These configurations lead to a $4$X (when $b=2$) and a $8$X (when $b=3$)
reduction in the sequence length. 
The Funnel-Transformer variants are pre-trained using the same task as in T5  on top of the C4 corpus~\cite{raffel2020exploring}.

\paragraph{Initial rankings.}
Since we primarily focus on the re-ranking setting, we consider several retrieval models to generate initial ranking candidates.
For the MS MARCO passage re-ranking task, we use BM25 (an implementation from Terrier~\cite{DBLP:conf/sigir/MacdonaldMSO12}) to generate the top-1K passages per query. In addition, we implemented 
the docT5query model~\cite{nogueira2019doc2query,nogueira2019docT5query} by training a T5 seq2seq model to generate 40 questions (i.e., expansions) per paragraph and use BM25 to retrieve top-1K passages. This serves as a high-recall initial ranking, wherein the recall@1K increases from 86.7 (MRR@10=19.3) in the base BM25 ranking to 93.76 (MRR@10=25.3) with document expansion. For the TREC DL Track, we use the official top-1k initial rankings from BM25~\cite{craswell2020overview, craswell2021overview}.

\paragraph{Efficiency metrics.}
To compare inference efficiency, we report FLOPs and latency as encouraged by \citet{dehghani2021efficiency}. To compute FLOPs we make use of a public repository~\footnote{\url{https://github.com/google-research/electra/blob/master/flops_computation.py}}.
To compute latency, we do as follows: each model is exported in the Tensorflow Saved Model format  before serving via the Tensorflow Model Server~\footnote{\url{https://www.tensorflow.org/tfx/tutorials/serving/rest_simple}} on a Intel Xeon CPU desktop with 8 CPU cores, 16 CPU threads, and 132 GB RAM.
We randomly select 500 queries and passages from the MS MARCO dataset. 
As for PreTTR~\cite{macavaney2020efficient},
to enable fair comparisons, 
we add an additional 500 queries, leading to a total of 1000 query-passages pairs,
to fully utilise the shared computation of the query encoder.
For each query-passage pair, we time the inference call to the model server 10 times and record the minimum. 
For each model, we report the 50 and 95-percentile of the 500 timing (1000 for PreTTR) as a two-number summary of latency. 
The time for tokenization is included for all models.
For PreTTR and ED2LM, we assume the token representations
of passages have already been loaded in the memory
akin to~\cite{macavaney2020efficient, gao2021coil}.

\section{Results}\label{sec.rerank_results}

In this section, we examine the effectiveness-efficiency trade-off of ED2LM
on the passage re-ranking task. 
The results of T5, ED2LM, BERT, and PreTTR have been displayed in Table~\ref{tab.rerank}. 
In Table~\ref{tab.compare_rerank},
we further summarise the comparisons (ED2LM vs. baseline models) from Table~\ref{tab.rerank} and highlight
the results that ED2LM provides a better trade-off.
We also visualise the results from different models
on the MS MARCO benchmark in Fig.~\ref{fig.msmarco_bm25_latency} when using docT5query~\cite{nogueira2019docT5query} as the initial ranking.

\subsection{Trade-off in Re-ranking}\label{sec.res_reranking}
\begin{table*}[!t]
    \centering
    \resizebox{1.01\textwidth}{!}{
    \begin{tabular}{c|cc|cc|cc||ccc}
    \toprule
         \multirow{2}{*}{Models} & \multicolumn{2}{c|}{MS MARCO (MRR@10)} &\multicolumn{2}{c}{Trec DL Track 2019}&\multicolumn{2}{|c||}{Trec DL Track 2020}& FLOPs&\multicolumn{2}{c}{Latency (ms)}\\
        & BM25+ & docT5query+& nDCG@10 & MAP& nDCG@10 & MAP &(T)&P50&P95\\
    \midrule
    
    \multicolumn{10}{c}{Baseline Models}\\
     \midrule
PreTTR ($p$)	&	36.7	&	37.4	&	70.0	&	39.8 	&	71.5 	&	45.5 	&	26	&	159	&	189	\\
BERT-base ($b$)	&	36.5	&	37.2	&	68.5	&	41.9 	&	71.9 	&	45.7	&	52	&	309	&	443	\\
T5-small ($t5s$)	&	35.9	&	36.6	&	68.8	&	42.3	&	68.1 &	42.1	&	22	&	123	&	127	\\
T5-base ($t5b$)	&	38.3	&	39.2	&	71.1	&	43.1 	&	73.7 	&	48.6 	&	67	&	405	&	425	\\
T5-large ($t5l$)	&	39.4	&	40.3	&	72.0 	&	42.9 	&	73.0 		&	48.0 	&	202	&	1111	&	1140	\\
T5-xl	($t5x$) &	39.6	&	40.6	&	71.8	&	42.2	&	74.6 	&	49.2	&	752	&	2490	&	2515	\\
\midrule
\multicolumn{10}{c}{Variants of ED2LM}\\
\midrule
ED2LM-small	&	37.2 ($\uparrow_{t5s}\downarrow_{t5blx}\uparrow_{b}$)	&	37.9 ($\uparrow_{t5s}\downarrow_{t5blx}\uparrow_{b}$)	&	69.5 ($\downarrow_{t5l}$)	&	40.8	&	69.6 ($\downarrow_{t5blx}$)	&	43.3 ($\downarrow_{t5blx}\downarrow_{b}$)	&	5	&	60	&	65	\\
ED2LM-base	&	38.7  ($\uparrow_{t5s}\downarrow_{t5lx}\uparrow_{b}\uparrow_{p}$)	&	39.6 ($\uparrow_{t5s}\downarrow_{t5lx}\uparrow_{b}\uparrow_{p}$)	&	70.2	&	42.5 ($\uparrow_{p}$)	&	71.5 ($\uparrow_{t5s}\downarrow_{t5x}$)	&	47.2 ($\uparrow_{t5s}\downarrow_{t5x}$)	&	21	&	157	&	185	\\
ED2LM-large	&	38.0 ($\uparrow_{t5s}\downarrow_{t5lx}\uparrow_{b}\uparrow_{p}$)	&	39.0 ($\uparrow_{t5s}\downarrow_{t5lx}\uparrow_{b}\uparrow_{p}$)	&	70.3	&	42.3 ($\uparrow_{p}$)	&	72.8 ($\uparrow_{t5s}$)	&	47.6 ($\uparrow_{t5s}$)	&	73	&	317	&	336	\\
ED2LM-xl	&	39.4 ($\uparrow_{t5sb}\uparrow_{b}\uparrow_{p}$)	&	40.4 ($\uparrow_{t5sb}\uparrow_{b}\uparrow_{p}$)	&	71.4	&	44.8 ($\uparrow_{t5sbx}\uparrow_{b}\uparrow_{p}$)	&	71.6 ($\uparrow_{t5s}\downarrow_{t5x}$)	&	48.2 ($\uparrow_{t5s}\uparrow_{b}\uparrow_{p}$)	&	287	&	811	&	834	\\
\midrule
\multicolumn{10}{c}{ED2LM with Funnel Blocks}\\
\midrule
ED2LM-F-$6L^{\times2}$	&	36.5 ($\downarrow_{t5blx}$)	&	37.4 ($\uparrow_{t5s}\downarrow_{t5blx}$)	&	68.0 ($\downarrow_{t5blx}$)	&	40.5 ($\downarrow_{t5b}$)	&	70.4 ($\downarrow_{t5bx}$)	&	44.1 ($\downarrow_{t5blx}$)	&	9	&	130	&	151	\\
ED2LM-F-$8L^{\times3}$	&	35.4 ($\downarrow_{t5blx}\downarrow_{b}\downarrow_{p}$)	&	36.2 ($\downarrow_{t5blx}\downarrow_{b}\downarrow_{p}$)	&	69.2 ($\downarrow_{t5l}$)	&	40.2 ($\downarrow_{t5bl}$)	&	70.5 ($\downarrow_{t5bx}$)	&	44.7 ($\downarrow_{t5blx}$)	&	7	&	108	&	126	\\

    \bottomrule
    \end{tabular}}
    \caption{The re-ranking performance when re-ranking top-1K paragraphs.
    We note down the significant difference at 
    0.05 level with $\uparrow$ and $\downarrow$ for the variants of ED2LM. 
    The comparisons are relative to T5-small, T5-base, T5-large, and, T5-xl (with subscriptions $t5s$, $t5b$, $t5l$, $t5x$),
    BERT-base (with subscriptions $b$), PreTTR with six layers of decoupled encoding (with subscriptions $p$).
    }
    \label{tab.rerank}
\end{table*}

\begin{figure*}[h!]
    \centering
    \includegraphics[width=18cm, trim=3.5cm 0cm 0cm 0cm,clip]{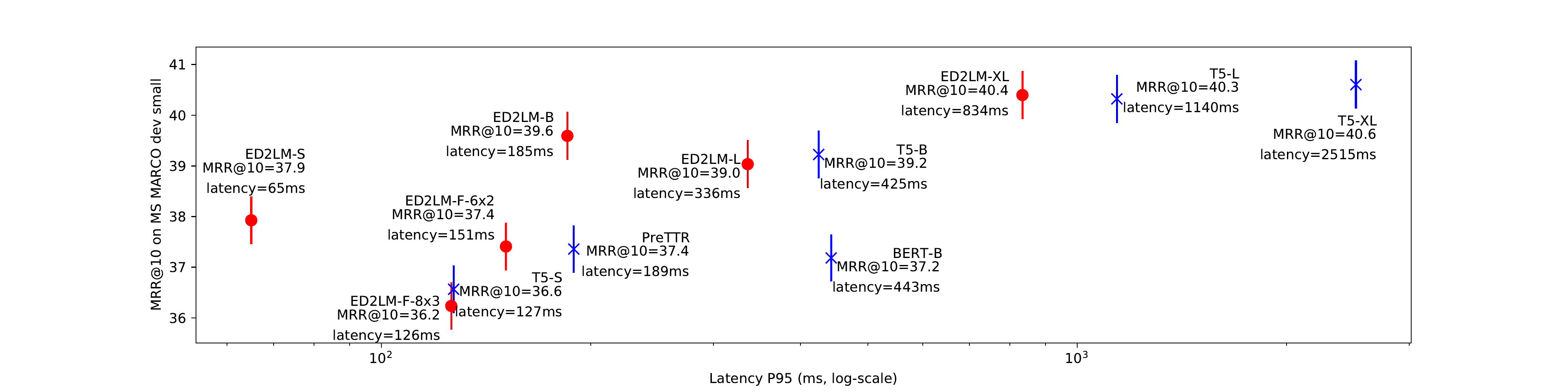}
    \caption{MRR@10 on MS MARCO dev small (6980 test queries) after re-ranking top-1K documents from docT5query~\cite{nogueira2019docT5query} vs. latency. The x-axis is the latency (95 percentile out of 500 calls); y-axis is the MRR@10 score. The point (ED2LM models) and the cross (baseline models) are the mean MRR@10 and the bar indicates the 95\% confidence interval.}
    \label{fig.msmarco_bm25_latency}
\end{figure*}

\paragraph{Results for the baseline models.} 
We achieve comparable results as previous studies on all three benchmarks.
In particular, \citep{nogueira2020T5ranking} reports $MRR@10=37.2$, $38.1$, $39.3$, and $39.8$
when using BERT-large, T5-base, T5-large, and T5-xl to re-rank
top-1K paragraphs from BM25 on MS MARCO passage re-ranking benchmark.
Besides, we list the re-ranking results on MS MARCO from 
COIL~\cite{gao2021coil} (MRR@10=34.8) and ColBERT~\cite{khattab2020colbert} (MRR@10=34.9) here for references.
For the TREC DL Track,
we select the submitted runs that are most comparable to ours, namely,
the top re-ranking run~\cite{yan2019idst} in 2019 (nDCG$@10=72.5$ and MAP$=45.3$)
and the 4th best re-ranking run~\cite{cao2020pingan}\footnote{The 1st-3rd best runs~\cite{qiaopash} in 2020 used TREC DL 2019 data
for fine-tuning.} for 2020 (nDCG$@10=73.7$ and MAP$=48.8$).

\paragraph{Effectiveness-efficiency trade-off.}
\begin{table*}[t!]
    \centering
    \resizebox{0.95\textwidth}{!}{
    \begin{tabular}{c|cccc||cc}
    \toprule
         ED2LM$\rightarrow$& Small & Base & Large & xl&F-$6L^{\times2}$&F-$8L^{\times2}$\\
         \midrule
        \multirow{2}{*}{T5-small} & F:4.4x/L:2.0x & \multirow{2}{*}{-}&\multirow{2}{*}{-}&\multirow{2}{*}{-}&F:2.4x/L:0.8x&\textbf{F:3.1x/L:1.0x}\\ 
         & r:$\uparrow$/n:\url{~}/m:\url{~} & &&& r:\url{~}/n:\url{~}/m:\url{~}&\textbf{r:\url{~}/n:\url{~}/m:\url{~}}\\
         \midrule
         \multirow{2}{*}{PreTTR} & F:5.2x/L:2.9x & F:1.2x/L:1.0x&\multirow{2}{*}{-}&\multirow{2}{*}{-}&F:2.9x/L:1.3x&F:3.7x/L:1.5x\\ 
        
         & r:\url{~}/n:\url{~}/m:\url{~} & r:$\uparrow$/n:\url{~}/m:$\uparrow$&&&r:\url{~}/n:\url{~}/m:\url{~}&r:$\downarrow$/n:\url{~}/m:\url{~}\\
         \midrule
        \multirow{2}{*}{BERT-base} & \textbf{F:10.4x/L:6.8x}& F:2.5x/L:2.4x&\multirow{2}{*}{-}&\multirow{2}{*}{-}&\textbf{F:5.8x/L:2.9x}&F:7.4x/L:3.5x\\ 
         &\textbf{r:$\uparrow$/n:\url{~}/m:$\downarrow$}&r:$\uparrow$/n:\url{~}/m:\url{~}&&&\textbf{r:\url{~}/n:\url{~}/m:\url{~}}&r:$\downarrow$/n:\url{~}/m:\url{~}\\
         \midrule
        \multirow{2}{*}{T5-base} &\multirow{2}{*}{-}& \textbf{F:3.2x/L:2.3x}&\multirow{2}{*}{-}&\multirow{2}{*}{-}&\multirow{2}{*}{-}&\multirow{2}{*}{-} \\ 
         &&\textbf{r:\url{~}/n:\url{~}/m:\url{~}}&&&&\\
         \midrule
        \multirow{2}{*}{T5-large} &\multirow{2}{*}{-}&F:9.6x/L:6.2x&F:2.8x/L:3.4x&\multirow{2}{*}{-}&\multirow{2}{*}{-}&\multirow{2}{*}{-} \\ 
         &&r:$\downarrow$/n:\url{~}/m:\url{~}&r:$\downarrow$/n:\url{~}/m:\url{~}&&&\\
         \midrule
        \multirow{2}{*}{T5-xl} &\multirow{2}{*}{-}&\multirow{2}{*}{-}&F:10.3x/L:7.5x&F:2.6x/L:3.0x&\multirow{2}{*}{-}&\multirow{2}{*}{-}\\ 
         &&&r:$\downarrow$/n:\url{~}/m:\url{~} &r:\url{~}/n:$\downarrow$/m:\url{~}&&\\
         \bottomrule
    \end{tabular}
    }
    \caption{The comparison of the effectiveness-efficiency trade-off for ED2LM derived from Table~\ref{tab.rerank}. Each row includes one baseline model,
    and individual columns are one of the ED2LM variants. 
    In each comparison (cell), the upper part is the efficiency
    comparison, where F indicates FLOPs and L is the latency (P95). In the lower part, the comparisons for the effectiveness are summarised. $\uparrow$, $\downarrow$, and, \url{~} denote the significant better, worse, and, no significant difference (at level 0.05) when comparing ED2LM models with the baseline. Herein, $r$ indicates MRR@10 on MS Marco dev small dataset (re-ranking top-1k from BM25); 
    $n$ and $m$ denote nDCG@10 and MAP, respectively, on TREC DL Track.
    We highlight comparisons that ED2LM could provide better effectiveness (MRR@10 or nDCG@10) or smaller latency.
    }
    \label{tab.compare_rerank}
\end{table*}
ED2LM decouples the encoding of the document and query, thereby allowing 
for caching the document representation offline. 
After pre-computing the document presentation as in PreTTR~\cite{macavaney2020efficient},
ED2LM achieves a highly favorable trade-off.
From Table~\ref{tab.rerank} and~\ref{tab.compare_rerank},
we make the following observations. (1) ED2LM-small and ED2LM-base perform at least as good as 
T5-small and T5-base, respectively, while providing more than a 2X speed up.
For ED2LM-base, its effectiveness is not significantly different from T5-large on both TREC DL Tracks
and under-performs by 0.7 (38.7 vs 39.4) on MS MARCO, while providing a 6.2X speed up.
When comparing with BERT-base and PreTTR, 
both ED2LM-small and ED2LM-base perform at least as good (for MRR@10 and nDCG@10) and are up to 6.8X faster. 
(2) ED2LM-large performs on par with T5-large on the TREC DL Tracks, but under performs on
MS MARCO by 1.4; whereas ED2LM-xl achieves similar MRR@10 on MS MARCO (39.4 vs 39.6),
but performs worse in terms of nDCG@10 on TREC DL Track 2020. 
Furthermore, in Fig.~\ref{fig.msmarco_bm25_latency} (MRR@10 on MS MARCO vs the latency (P95) by re-ranking the top-1K from docT5query)
the leftmost ED2LM-small achieves better effectiveness than
T5-small, PreTTR, and BERT-base. Likewise,  ED2LM-base achieves similar latency as PreTTR and 
is 2.3X more efficient than BERT-base but 
achieves higher MRR@10.
In the meantime, though more efficient, 
ED2LM-xl and ED2LM-large perform  close to their counterparts, once again confirming the observations.
We argue that, 
on the one hand,  co-training of query likelihood 
and the discriminative cross-entropy leads to better ranking quality, 
which is especially true for the smaller variants (small and base);
On the other hand, not attending to the query during document encoding
leads to performance decreases, which dominates the outcomes 
in larger model variants (like large and xl).

\paragraph{ED2LM-F: Storage compression with Funnel Transformer.}
The results for the two variants of ED2LM with Funnel blocks are summarised 
in the bottom block of Table~\ref{tab.rerank} and the rightmost columns in Table~\ref{tab.compare_rerank}.
In terms of storage, ED2LM-F-$6L^{\times2}$ provides 
$4$X compression and ED2LM-F-$8L^{\times3}$
provides $8$X compression by reducing the sequence length
in the encoder.
It can be seen that, ED2LM-F-$6L^{\times2}$
outperforms T5-small and 
performs as well as BERT-base and PreTTR.
Furthermore, while ED2LM-F-$8L^{\times3}$ provides 8X compression, 
the effectiveness drops below that of T5-small and BERT-base on the MS MARCO benchmark. However, it achieves on-par results relative to 
T5-small and BERT-base on the TREC DL Track in terms of both nDCG@10 and MAP.
As for efficiency, ED2LM-F-$8L^{\times3}$ is similar to T5-small and PreTTR, but is 3.5X faster than BERT-base.

\subsection{Ablation Analysis}\label{sec.analysis}

\begin{table*}[!t]
    \centering
    \resizebox{0.6\textwidth}{!}{
    \begin{tabular}{c|cc|c}
    \toprule
    Models & \multicolumn{2}{c|}{MS Marco}\\
    &Training Data &Loss & MRR@10 \\
        \midrule
    PreTTR  & MS Marco &-&35.2\\
    T5-base & MS Marco &- &  38.4 \\
     T5-base & RQA-Hard&-&  38.0 \\
    ED2LM-base & MS Marco&-& 37.5 \\
    ED2LM-base & RQA-Hard&-&  37.3\\
    \midrule 
    ED2LM-base & MS Marco &LUL~\cite{dos2020beyond} & 31.2\\
    ED2LM-base & RQA-Merge &LUL~\cite{dos2020beyond} & 33.6\\
    ED2LM-base & RQA-Merge &MLE (Eq.~\ref{eq.loss_ql})&  30.2\\
    ED2LM-base & RQA-Merge &CE (Eq.~\ref{eq.loss_disc})&  38.2\\
    \bottomrule
    \end{tabular}}
    \caption{Ablation study.
    In the upper half, the uses of alternative training data are explored.
    In the lower half, different loss functions are used to train ED2LM, including the LUL loss from \cite{dos2020beyond}, negative log-likelihood loss on questions as in~\cite{nogueira2019docT5query}, and the cross-entropy loss on true/false token as in~\cite{nogueira2020T5ranking}.}
    \label{tab.ablation}
\end{table*}

\paragraph{The use of RocketQA-Merge dataset for training.}
In our experiments, we find that the ranking quality of 
the proposed ED2LM, as well as PreTTR model, benefit considerably
from RocketQA-Merge.
We demonstrate the training performance (upper part) in Table~\ref{tab.ablation} on RocketQA and the MS MARCO training dataset.
It can be seen that 
T5 achieves similar performance on both training data sets. 
In the meantime,
ED2LM achieves MRR@10=37.5 when trained on the MS MARCO training dataset, and can achieve 38.7 when trained on the ``RQA-Merge'' dataset. 
This is also true for PreTTR, which sees an MRR@10  increase from 35.2 to 36.7.
We conjecture that the decoupled encoding of query and documents, as in ED2LM and PreTTR, 
requires more queries for training whereas models that use full cross-attention benefit less from the extra training data.
The training performance of ED2LM-base on RocketQA-Hard in Table~\ref{tab.ablation} provides evidences for this, where
ED2LM-base achieves an even lower MRR@10.
RocketQA-Hard is a subset of RocketQA-Merge and includes hard negative samples but without the extra queries.
Therefore, we conclude that \textit{more unique questions for training is one of the ED2LM's
key ingredients.}

\paragraph{Alternative loss functions for training.}
In~\cite{dos2020beyond}, the unlikelihood loss (referred as LUL)
was used to train a BART~\cite{lewis2020bart} model for question answering.
In this section, we train ED2LM using the LUL loss from~\cite{dos2020beyond} on both the MS MARCO and RQA-Merge training sets. We also use the negative log-likelihood loss in Eq.~\ref{eq.loss_ql} (as in docT5query~\cite{nogueira2019docT5query}) and the cross-entropy loss in
Eq.~\ref{eq.loss_disc} (as in~\cite{nogueira2020T5ranking}) to train ED2LM separately. 
From Table~\ref{tab.ablation} (lower part), LUL leads to significantly worse MRR@10 
than using the loss in Eq.~\ref{eq.loss} (33.6 vs 38.7), but outperforms the
use of negative log-likelihood loss from Eq.~\ref{eq.loss_ql} as in~\cite{zhuang2021deep}.
When only using the cross-entropy loss of the true/false token (Eq.~\ref{eq.loss_disc}), 
effectiveness is slightly worse than when using the loss in combination with query likelihood (38.2 vs 38.7), mirroring the findings from~\cite{ju2021text}.
Therefore, we conclude that \textit{the use of both true/false tokens and query likelihood for training (as in Eq.~\ref{eq.loss}) 
is another key ingredient for ED2LM.}

\section{ED2LM for Question Generation}\label{sec.qgen_results}
\begin{table*}[!t]
    \centering
    \resizebox{0.8\textwidth}{!}{
    \begin{tabular}{c|l|c}
    \toprule
    \multicolumn{3}{c}{Paragraph} \\
    \midrule
    \multicolumn{3}{l}{An experience modifier is an adjustment factor assigned to an Employer's FEIN by the rating bureau} \\
    \multicolumn{3}{l}{(NCCI or State Bureau). The factor compares your loss data to other employers with the same class} \\
    \multicolumn{3}{l}{codes, and is expressed as a credit or debit on your policy.} \\
    \midrule
    Model & \multicolumn{1}{c|}{Question} & Answerable ?\\
    \midrule
    T5 & is a modifier factor English & No \\
    T5 & what is experience modifier rating & No \\
    ED2LM & what is an experience modifier in an insurance policy & Yes \\
    ED2LM & experience modifier definition & Yes \\
    \bottomrule
    \end{tabular}}
    \caption{Example generations from ED2LM-base and T5-base.}
    \label{tab.gensample}
\end{table*}

Question generation has played an important role for different downstream tasks~\cite{shakeri-etal-2020-end,puri-etal-2020-training,del2021question}.
We conjecture that the combination of generation and ranking loss used in ED2LM has the potential to improve question generation when compared to models trained with generation loss only. We evaluate this conjecture by comparing questions generated by vanilla generator trained with question likelihood
only~\cite{nogueira2019doc2query} and ED2LM in different scenarios:
manual inspection, assessment with automatic metrics and synthetic training data generation. For question generation task, an extra ``eos'' token (namely, the end of sequence token) is inserted between the question and the true/false token. Our pilot experiments show that this change does not influence the ranking performance but boosts the generation quality of ED2LM. We adopt
the top-k sampling decoding~\cite{fan2018hierarchical} (set $k=10$) in question generation for all models.

\subsection{Question Generation with Less Hallucination}

\paragraph{Manual inspection of the generated questions.}
We investigate the reasons why ED2LM can significantly outperform deep query likelihood 
(MRR@10=38.7 vs 30.2 from Table~\ref{tab.ablation}) by a big margin.
We compare the questions generated by ED2LM and T5 trained with query likelihood as in Eq.~\ref{eq.loss_ql}.
We sample 66 documents from the MS MARCO passage corpus with at least one correct 
query in the MS MARCO development dataset, and collect 10 unique generated queries from both ED2LM and T5,
ending up with 660 query-documents pairs for annotation.
These pairs are labeled by eight annotators with a single binary question: ``Is the generated query (question) answered by the given document (passage)?''. 
We avoid potential bias during annotation by not informing the annotators which system generated which questions.
According to the annotated data, \textbf{70.6\%} of the queries generated by ED2LM are answerable by the source document, while \textbf{52.1\%} of the queries generated by T5 are answerable.
We conjecture that 
\textit{the use of Eq.~\ref{eq.loss} for training makes the query generator
stick to the document better, leading to fewer hallucinations, thus producing
better ranking when the decoder is used as a ranker.}

\paragraph{Question vs. paragraph overlap.} 
We measured the overlap between generated questions and their respective source passages using a set of 3k generated questions from each system.
Intuitively, question generators that hallucinate less are more likely to stick to the text from the source paragraph.
The overlap is computed as the macro-average of the question-paragraph word-level overlap, and is normalised using the length of the question.
While T5 has an overlap rate of \textbf{55.62\%} (i.e., 55.62\% of question tokens also appear in the source paragraph), ED2LM has an overlap rate of \textbf{62.14\%}, which is more than 6\% higher than T5 model.
In Table~\ref{tab.gensample},
we present example questions generated by T5 and ED2LM and their source paragraphs.
Although T5 questions are somewhat related to the paragraph, the paragraph is not a good answer for them. 
For example, in Table~\ref{tab.gensample}, the first question T5 hallucinates the word \emph{English}, which  compromises the question quality.

\subsection{Synthetic Training Data for Retrieval} 
Finally, we demonstrate the advantages of the generated questions
from ED2LM by using them to train 
a dual-encoder based passage retrieval model, following the configurations in~\cite{lu2021npr}.
Specifically,
we train a BERT$_{large}$ dual encoder model using the synthetic question-passage pairs generated by ED2LM and T5 respectively and report the results on MS MARCO dev set. For each passage, we generate three synthetic questions. We also extract hard negatives by randomly sampling passages from the same document.
During training, we use both in-batch negatives and hard negatives. 
 During inference, we retrieve top 1000 passages for each question from the passage collection containing about 8.8 million passages and report MRR@10.
 The model using ED2LM generated data achieves \textbf{MRR@10=30.4}, whereas
 the model using T5 generated data gets \textbf{MRR@10=26.5}.
We argue that the boost is due to that the synthetic training data
from ED2LM is with less generation hallucination (18\% according to the manual annotation), thus including few training noise. 
\section{Conclusion}
 In this work, 
 we propose a novel model named ED2LM.
 ED2LM encodes documents and decodes the query using a trailing binary class token appended to the query for ranking.
 By training on a dataset with more unique questions (namely, ``RocketQA-Merge''~\cite{qu2021rocketqa}) and optimizing both query likelihood and a
 discriminative loss over the true/false token, 
ED2LM achieves competitive results 
compared to corresponding T5 models. 
When used as a
 decoder-only language model during inference, ED2LM provides up to 6.8X 
 speedup without sacrificing effectiveness.
We further demonstrate that ED2LM could generate questions with 
less hallucination. 
 For future works, 
 we plan to investigate the uses of ED2LM 
 for different (generation) tasks 
 such as multi-sentence compression (MRC)~\cite{zhao2019unsupervised},
 headline generation~\cite{shen2019select}, and list question answering~\cite{katti2021question}.

\newpage
\bibliography{ranking}
\bibliographystyle{acl_natbib}

\appendix
\section{Appendix}\label{sec.appendix}

\subsection{Re-ranking on Natural Questions}
\begin{table}[!h]
    \centering
    \resizebox{0.5\textwidth}{!}{
    \begin{tabular}{c|cccc}
    \toprule
        & Hits@1 & Hits@5 & Hits@10 &HitsS@20\\
         \midrule
         initial ranking & 52.47&72.24&77.7&81.33\\
         \midrule
T5-small&50.54&72.32&78.83&83.29\\
ED2LM-small&52.20 $\Uparrow$&72.65 &78.39 &82.27 $\Downarrow$\\
\midrule
T5-base&58.88&77.20&81.08&84.10\\
ED2LM-base&54.03 $\Downarrow$&76.06 $\Downarrow$&80.35 $\downarrow$&83.54 $\downarrow$\\
\midrule
T5-large&64.48&79.52&82.77&85.15\\
ED2LM-large&58.69 $\Downarrow$&76.86 $\Downarrow$&81.27 $\Downarrow$&84.10 $\Downarrow$\\
\midrule
T5-xl&65.64&80.19&83.46&85.51\\
ED2LM-xl&61.85 $\Downarrow$&78.91 $\Downarrow$&82.43 $\Downarrow$&84.70 $\Downarrow$\\
    \bottomrule
    \end{tabular}}
    \caption{The models are fine-tuned on NQ training data and validated on the NQ dev. We show the results on NQ test with 3610 questions, by re-ranking top-100 paragraphs (adv-hn) from DPR~\cite{karpukhin2020dense} for NQ.}
    \label{tab.nq_rerank}
\end{table}
We further examine the re-ranking performance of ED2LM on NQ dataset~\cite{kwiatkowski2019natural}.
The NQ dataset includes 79k user queries
 from the Google search engine. 
The subset of NQ derived in~\cite{karpukhin2020dense}
are used. The data has the form (question, passage, label), wherein only the queries with short answers are included.
The task is to retrieve and re-rank the chunked paragraphs
 from Wikipedia with up to 100 words for the queries.
The training data is balanced by up-sampling the positive training samples.
For validation purposes, we measure Hits@20 on the dev dataset with 8757 questions.
We report Hits@1, 5, 10, and, 20 by re-ranking the top-100 search results for the 3610 test questions
from DPR~\cite{karpukhin2020dense} (the run named adv-hn)\footnote{https://github.com/facebookresearch/DPR}.

The results are summarised in Table~\ref{tab.nq_rerank}. While ED2LM performs significantly worse than its T5 counterpart, 
it boosts the ranking quality considerably by re-ranking the initial ranking. 
In terms of Hits@1, ED2LM performs worse by almost 9\% (e.g., ED2LM-large) relative to T5 but still provides up to 18\% improvements (e.g., ED2LM-xl)
over the dual-encoder initial ranking. In terms of Hits@5, Hits@10, and Hits@20, 
though ED2LM still performs significantly worse, the gap becomes smaller, namely, mostly less than 2\% (apart from ED2LM-large performs worse by 3.35\%), when providing up to 9\% improvements (e.g., Hits@5 using ED2LM-xl) over the initial ranking. On the one hand, comparing with its performance on MS Marco, ED2LM performs worse on NQ relative to the T5 models, perhaps due to the relatively small amount of training data available in NQ ($\sim$50k unique queries vs $\sim$550k in MS Marco). On the other hand, the results on NQ actually double-confirm the promising ranking quality of the ED2LM model, especially considering its ability to provide ranking scores for the generate text (as discussed in Section~\ref{sec.qgen_results}).

\end{document}